\documentclass[10pt,twocolumn,letterpaper]{article}
\usepackage{times}
\usepackage{epsfig}
\usepackage{graphicx}
\usepackage{amsmath}
\usepackage{amssymb}
\usepackage{graphicx}
\graphicspath{{./pics/}}
\usepackage{subcaption}
\usepackage[usenames, dvipsnames]{color}
\usepackage{amsthm}
\usepackage{amsfonts}
\usepackage{amscd}
\usepackage{mathrsfs}
\usepackage{bm}
\usepackage{enumerate}
\usepackage{algorithmic, algorithm}
\usepackage{multirow}
\usepackage{authblk}
\newcommand{\R}{\mathbb{R}}
\newcommand{\M}{\mathcal{M}}

\newcommand{\N}{\mathcal{N}}

\newcommand{\bx}{\bm{x}}

\newcommand{\bp}{\bm{p}}
\newcommand{\bq}{\bm{q}}
\newcommand{\btheta}{\bm{\theta}}
\newcommand{\bxi}{\bm{\xi}}

\newtheorem{thm}{Theorem}
\newtheorem{remark}{Remark}
\usepackage{algorithmic, algorithm}
\usepackage{array}
\newcolumntype{C}[1]{>{\centering\let\newline\\\arraybackslash\hspace{0pt}}m{#1}}

\usepackage[pagebackref=true,breaklinks=true,letterpaper=true,colorlinks,bookmarks=false]{hyperref}

\begin{document}

\title{\textit{LDMNet}: Low Dimensional Manifold Regularized Neural Networks\footnote{Work partially supported by NSF, DoD, NIH, and Google.}}

\author[1]{Wei Zhu}
\author[2]{Qiang Qiu}
\author[3]{Jiaji Huang}
\author[2]{Robert Calderbank}
\author[2]{Guillermo Sapiro}
\author[1]{Ingrid Daubechies}
\affil[1]{Department of Mathematics, Duke University}
\affil[2]{ECE, Duke University}
\affil[3]{Baidu Silicon Valley AI Lab}

\date{}
\maketitle

\begin{abstract}
Deep neural networks have proved very successful on archetypal tasks for which large training sets are available, but when the training data are scarce, their performance suffers from overfitting. Many existing  methods of reducing overfitting are data-independent, and their efficacy is often limited when the training set is  very small. Data-dependent regularizations are mostly motivated by the observation that data of interest lie close to a manifold, which is typically hard to parametrize explicitly and often requires human input of tangent vectors. These methods typically only focus on the geometry of the input data, and do not necessarily encourage the networks to produce geometrically meaningful features. To resolve this, we propose a new framework, the Low-Dimensional-Manifold-regularized neural Network (\textit{LDMNet}), which incorporates a feature regularization method that focuses on the geometry of both the input data and the output features. In \textit{LDMNet}, we regularize the network by encouraging the combination of the input data and the output features to sample a collection of low dimensional manifolds, which are searched efficiently without explicit parametrization. To achieve this, we directly use the manifold dimension as a regularization term in a variational functional. The resulting Euler-Lagrange equation is a Laplace-Beltrami equation over a point cloud, which is solved by the point integral method without increasing the computational complexity. We demonstrate two benefits of \textit{LDMNet} in the experiments. First, we show that \textit{LDMNet}  significantly outperforms widely-used network regularizers such as weight decay and \textit{DropOut}. Second, we show that \textit{LDMNet} can be designed to extract common features of an object imaged via different modalities, which implies, in some imaging problems, \textit{LDMNet} is more likely to find the model that is subject to different illumination patterns. This proves to be very useful in  real-world applications such as cross-spectral face recognition.

\end{abstract}

\section{Introduction}
In this era of big data, deep neural networks (DNNs) have achieved great success in machine learning research and commercial applications. When large amounts of training data are available, the capacity of DNNs can  easily be increased by adding more units or layers to extract more effective high-level features \cite{Goodfellow-et-al-2016,He_2016_CVPR,Wan_2013_ICML}. However, big networks with millions of parameters can easily overfit even the largest of datasets. It is thus crucial to regularize DNNs so that they can extract ``meaningful'' features not only from the training data, but also from the test data.

Many widely-used network regularizations are data-independent. Such techniques include  weight decay, parameter sharing, \textit{DropOut} \cite{Srivastava_2014_dropout}, \textit{DropConnect} \cite{icml2013_wan13}, etc. Intuitively, weight decay alleviates overfitting by reducing the magnitude of the  weights and the features, and \textit{DropOut} and \textit{DropConnect} can be viewed as computationally inexpensive ways to train an exponentially large ensemble of DNNs. Their effectiveness as network regularizers can be quantified by analyzing the Rademacher complexity, which provides an upper bound for the generalization error \cite{bartlett1998sample,icml2013_wan13}.

\begin{figure*}
    \centering
    \begin{subfigure}[t]{0.22\textwidth}
        \includegraphics[width=\textwidth]{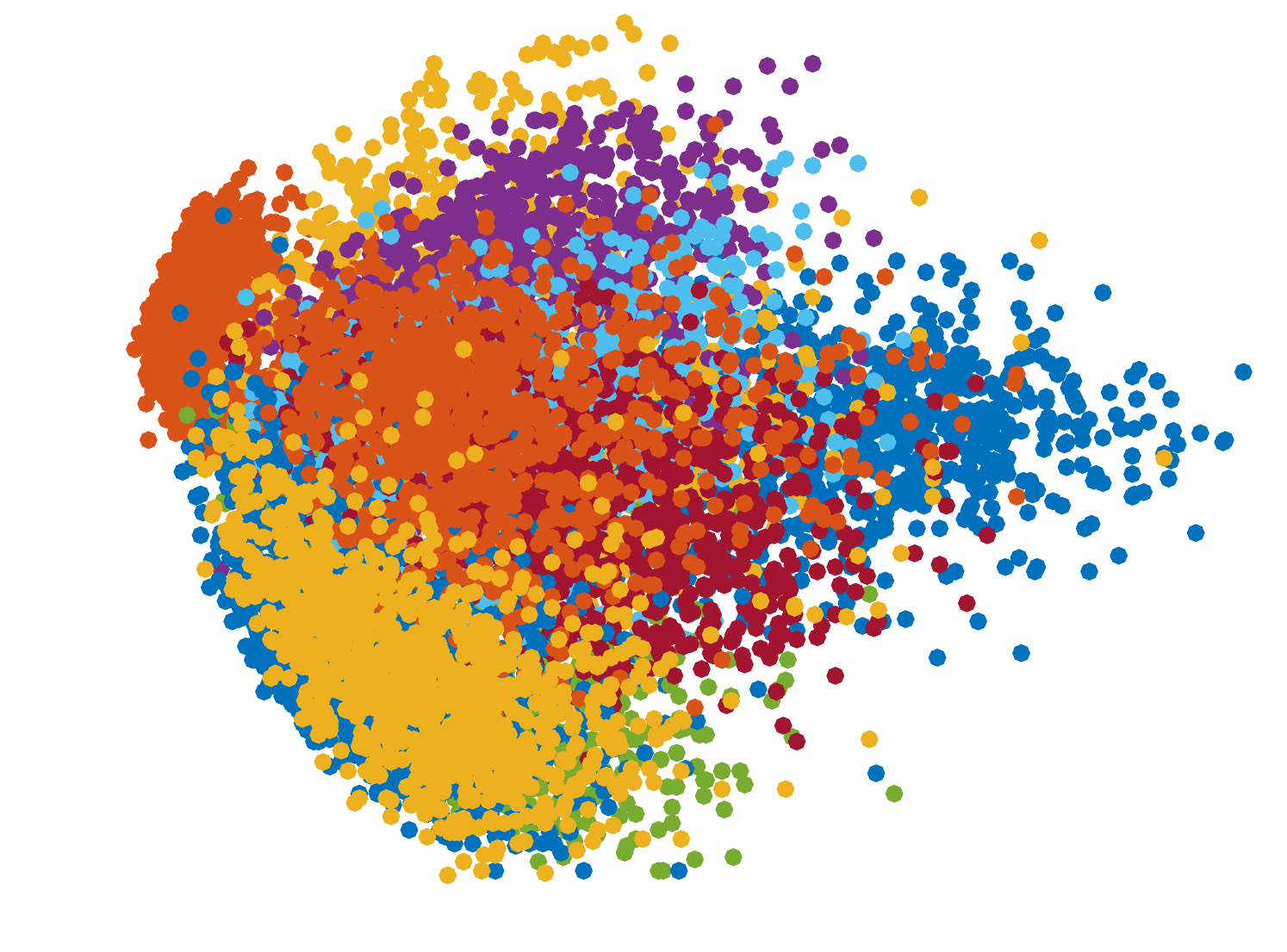}
        \caption{10,000 test data from MNIST}
        \label{fig:mnist_original}
    \end{subfigure}
    ~ 
    \begin{subfigure}[t]{0.22\textwidth}
        \includegraphics[width=\textwidth]{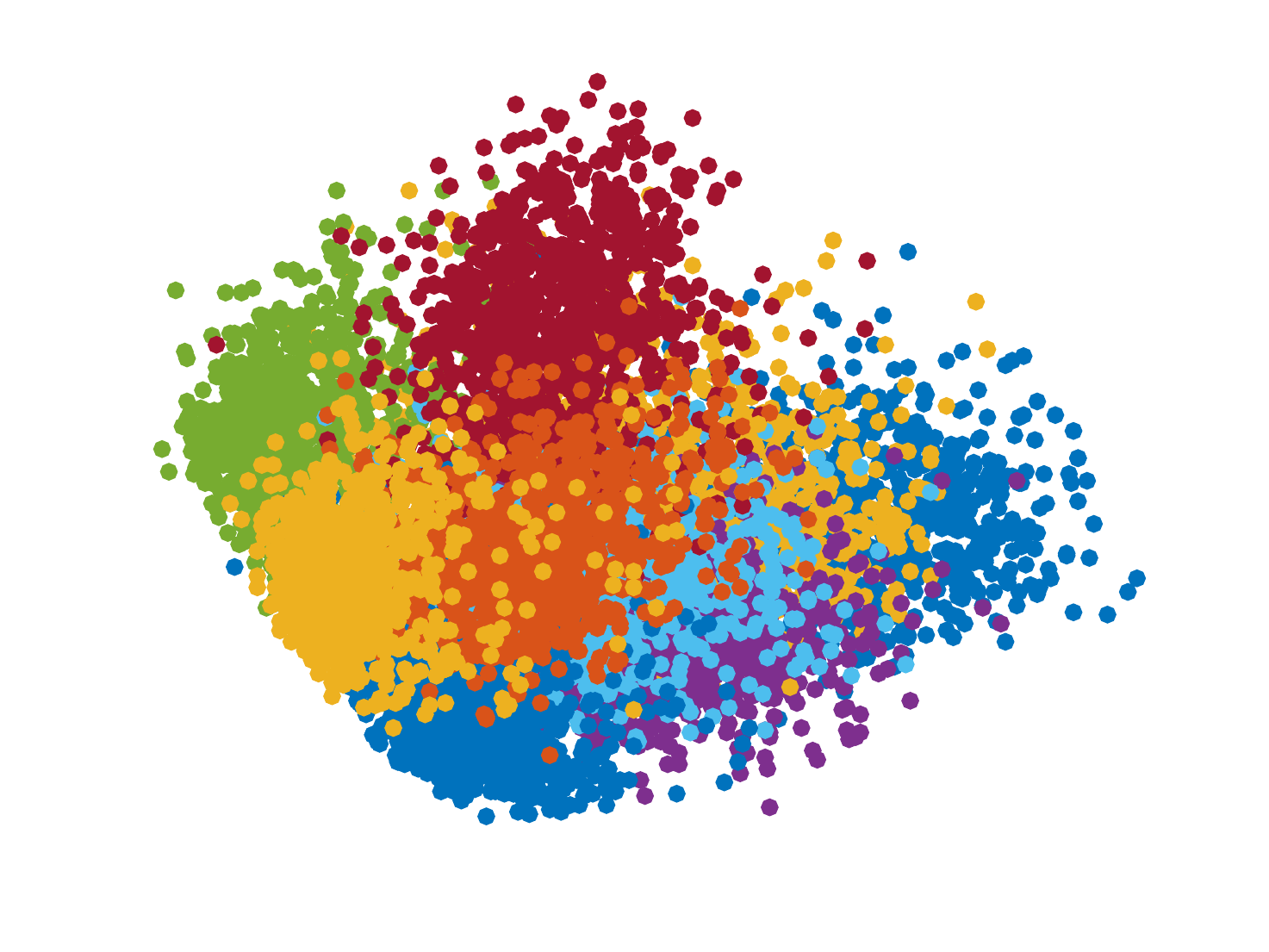}
        \caption{Features learned from a network regularized by weight decay}
        \label{fig:mnist_wd}
    \end{subfigure}
    ~ 
    \begin{subfigure}[t]{0.22\textwidth}
        \includegraphics[width=\textwidth]{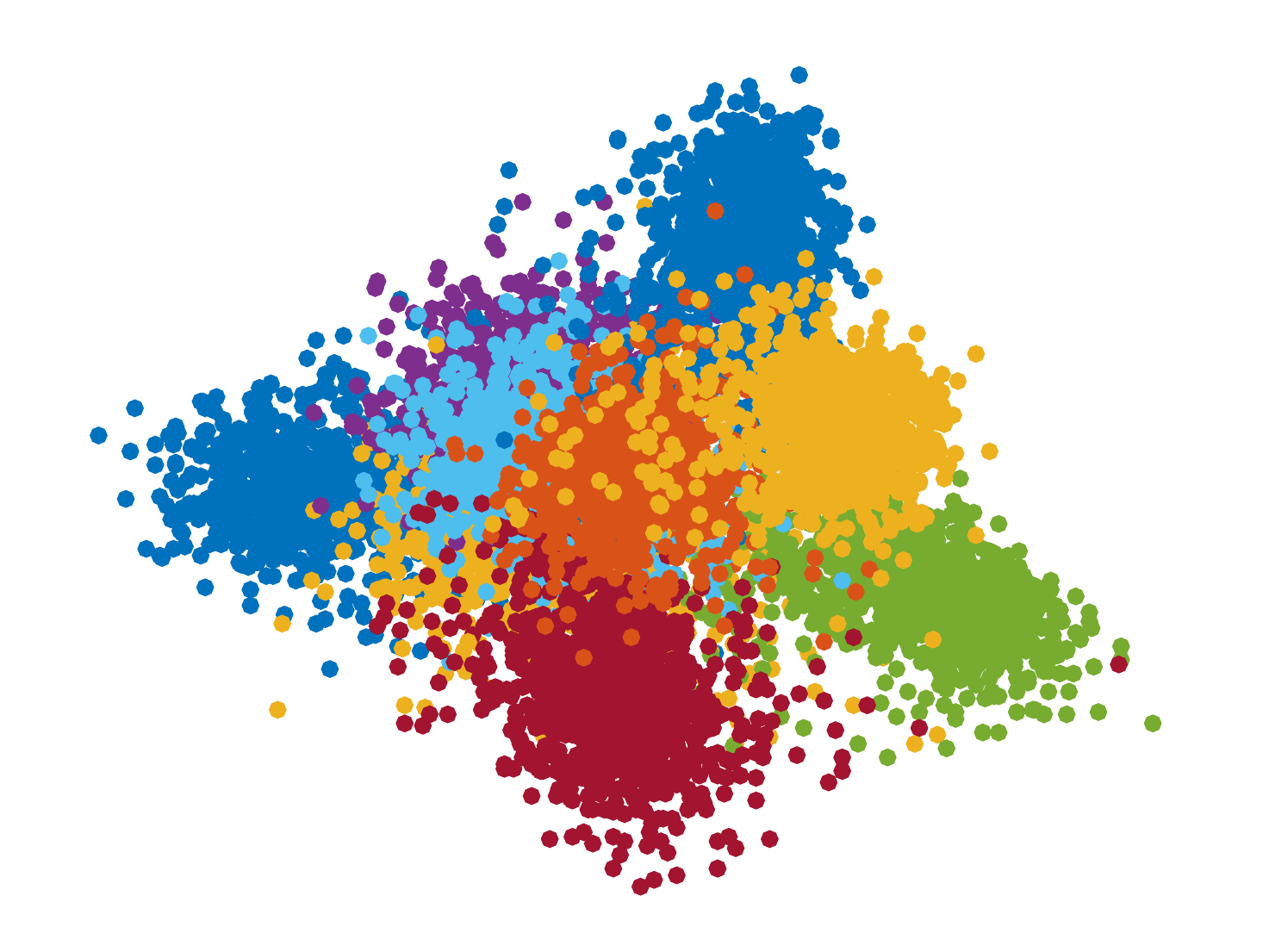}
        \caption{Features learned from a network regularized by \textit{DropOut}}
        \label{fig:mnist_dropout}
    \end{subfigure}
    ~ 
    \begin{subfigure}[t]{0.22\textwidth}
        \includegraphics[width=\textwidth]{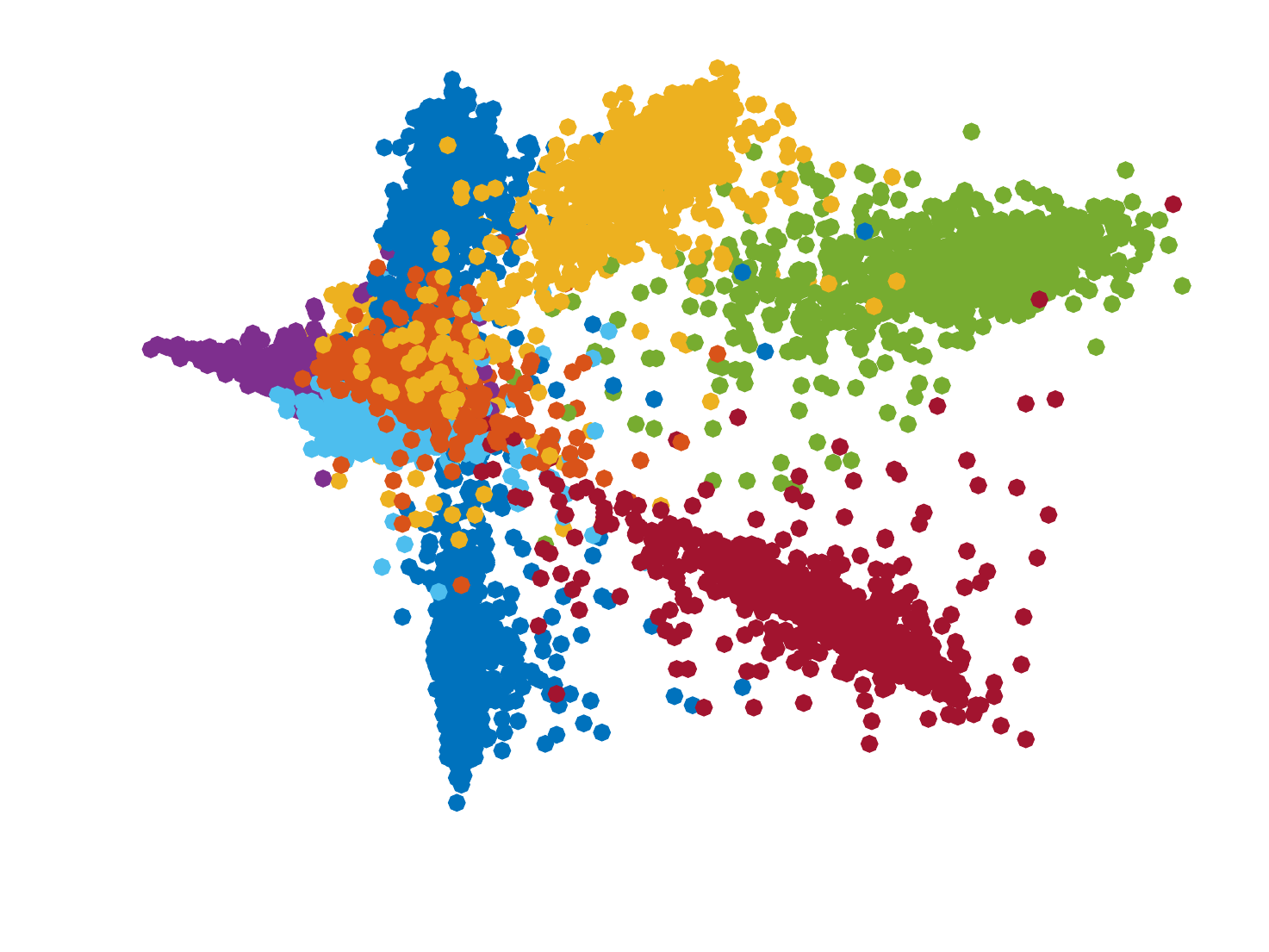}
        \caption{Features learned from a network regularized by \textit{LDMNet}}
        \label{fig:mnist_ldm}
    \end{subfigure}
    \caption{Test data of MNIST and their features learned by the same network with different regularizers. All networks are trained from the same set of 1,000 images. Data are visualized in two dimensions using PCA, and ten classes are distinguished by different colors. In (b) and (c), the features learned by weight decay and \textit{DropOut} typically sample two dimensioanl regions. Whereas in (d), features learned by \textit{LDMNet} tend to concentrate on one-dimensional and zero-dimensional manifolds (curves and points). }\label{fig:visual_mnist}
\end{figure*}

\begin{figure*}
    \centering
    \begin{subfigure}[b]{0.25\textwidth}
        \includegraphics[width=\textwidth]{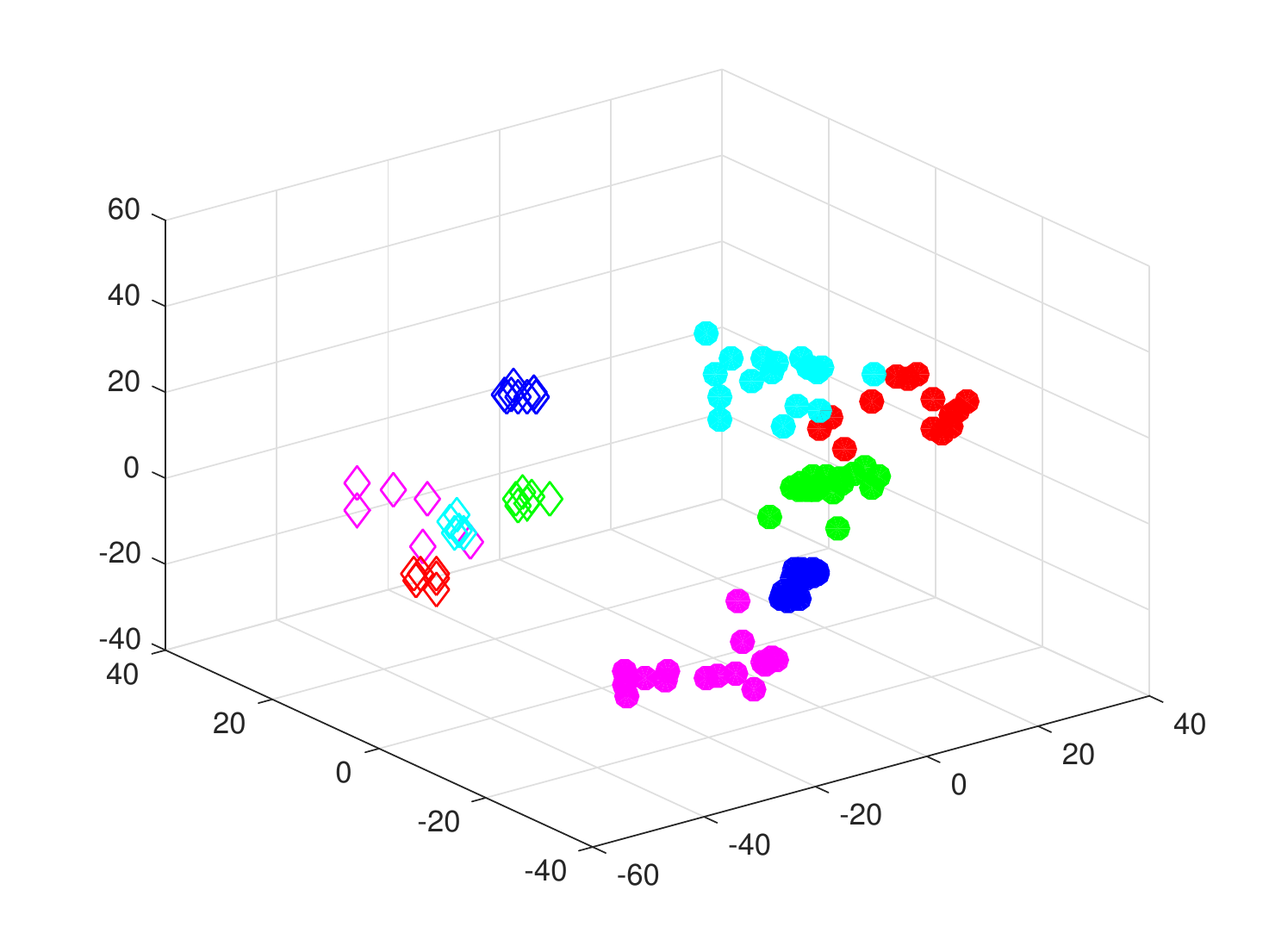}
        \caption{VGG-face}
        \label{fig:vgg}
    \end{subfigure}
    ~ 
    \hspace{-5mm}
    \begin{subfigure}[b]{0.25\textwidth}
        \includegraphics[width=\textwidth]{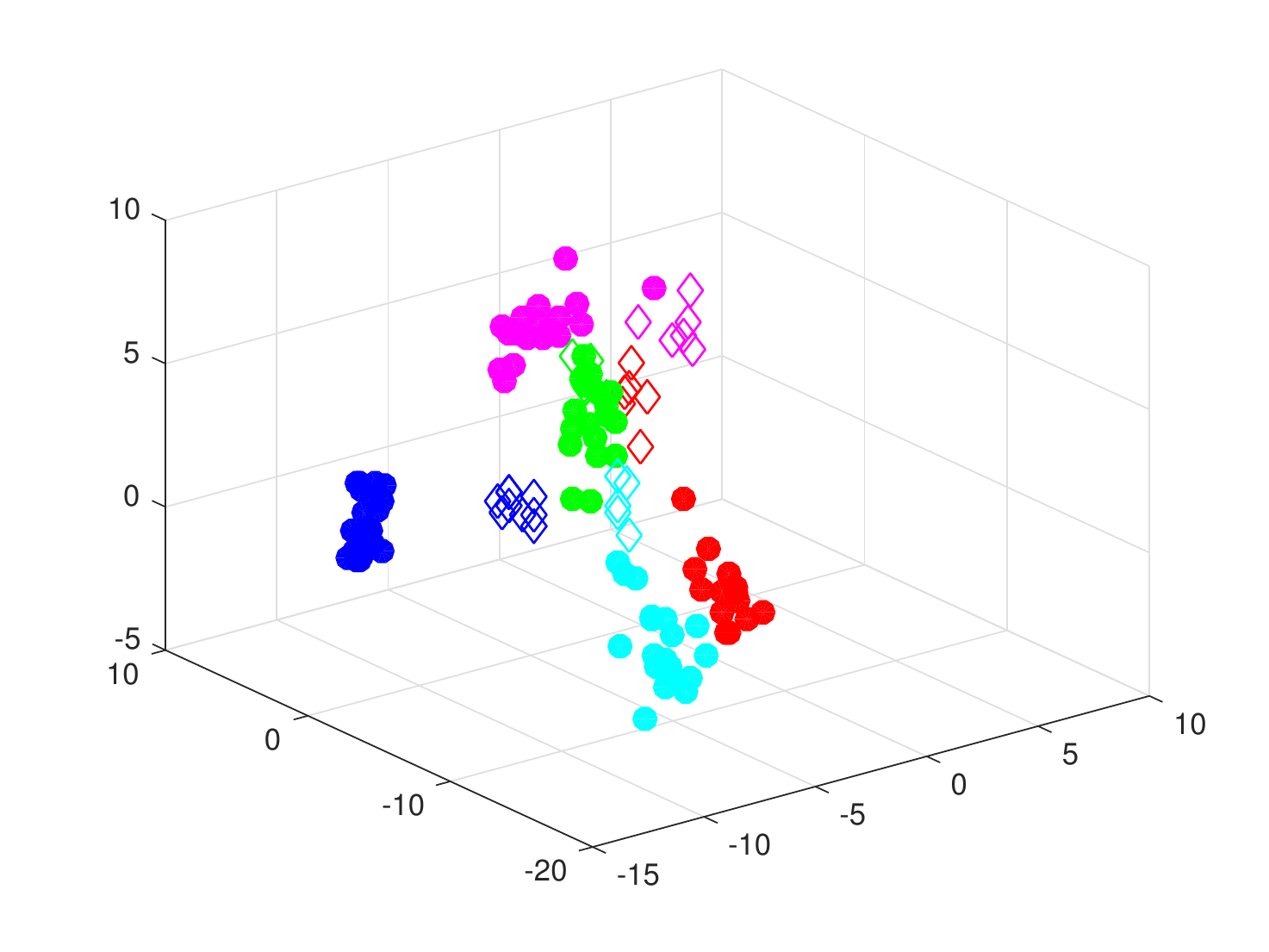}
        \caption{Weight decay}
        \label{fig:wd}
    \end{subfigure}
    ~ 
    \hspace{-5mm}
    \begin{subfigure}[b]{0.25\textwidth}
        \includegraphics[width=\textwidth]{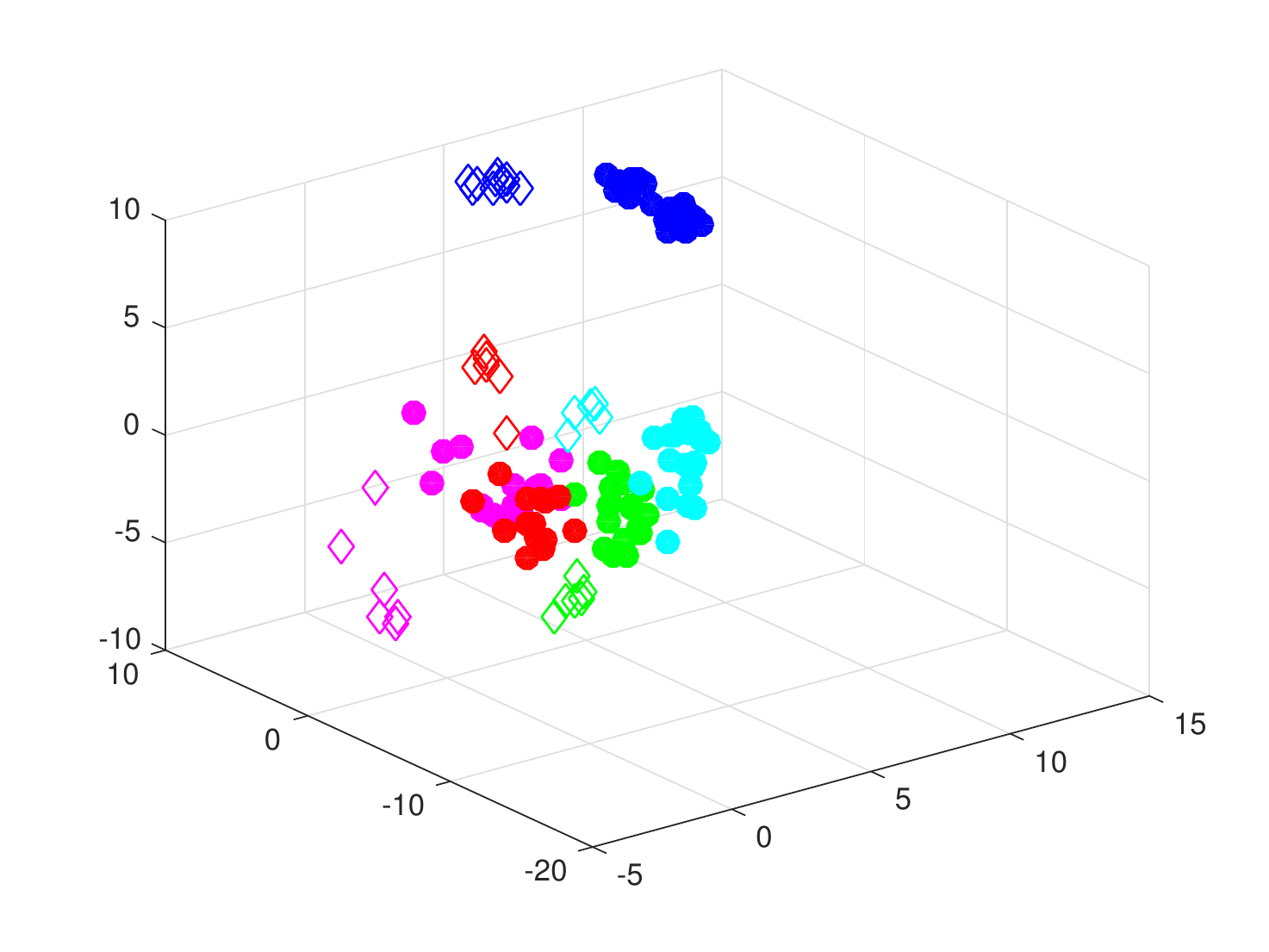}
        \caption{\textit{DropOut}}
        \label{fig:dropout}
    \end{subfigure}
    ~ 
    \hspace{-5mm}
    \begin{subfigure}[b]{0.25\textwidth}
        \includegraphics[width=\textwidth]{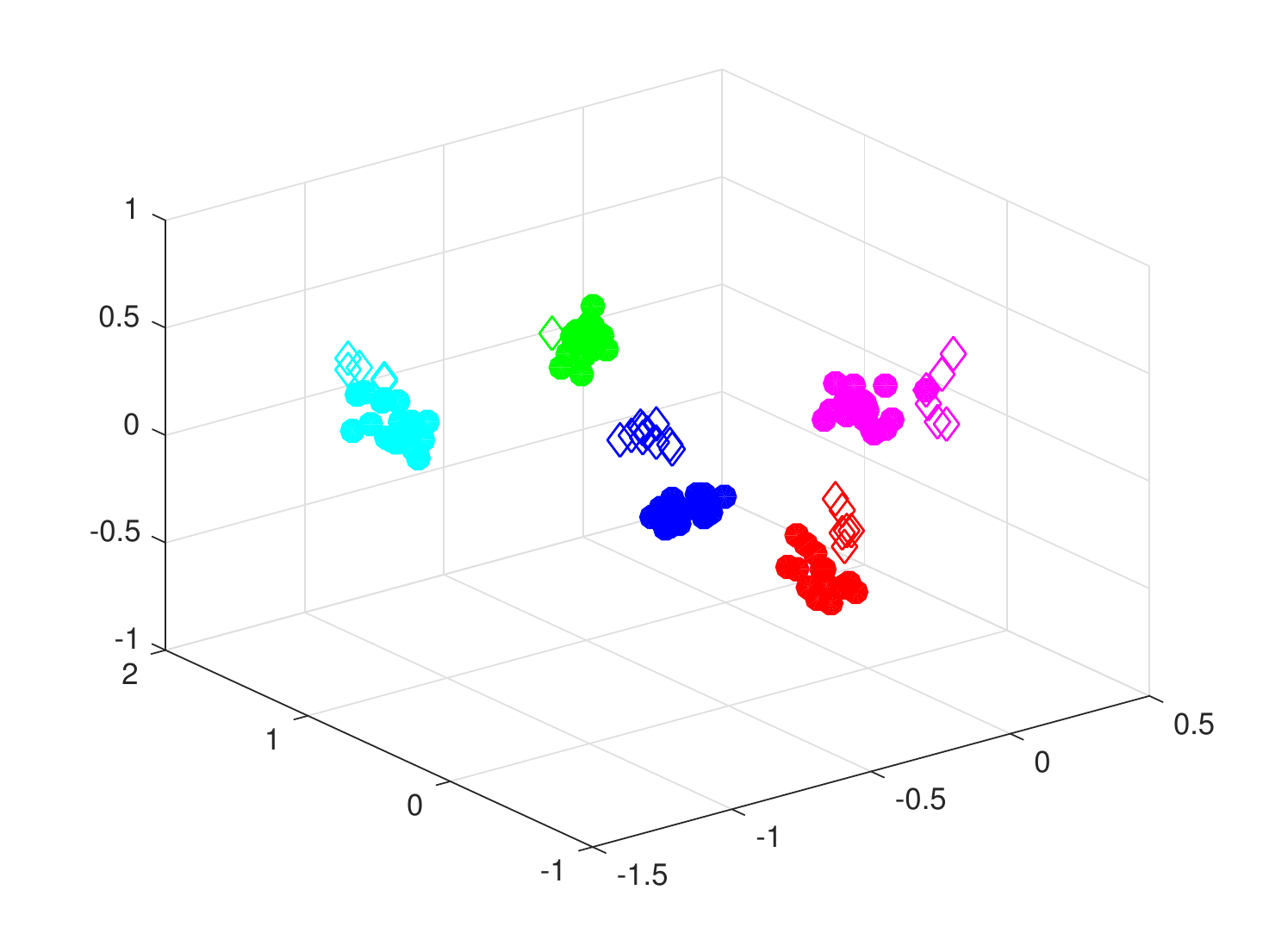}
        \caption{\textit{LDMNet}}
        \label{fig:ldm}
    \end{subfigure}
    \caption{Features generated from VGG-face model \cite{parkhi2015deep} with or without embedding learned from the network specified in Table~\ref{tab:nirvis_net} with different regularizations. A total of five subjects are displayed, with one color per subject. VIS images are denoted as filled circles and NIR images are denoted as unfilled diamonds. All features are visualized in two dimensions using PCA. In (a), (b), and (c), features of the same subject typically form two clusters, one for NIR and the other for VIS. In (d), with \textit{LDMNet}, features of the same subject from  different modalities merge to form a single low dimensional manifold.}\label{fig:visual}
\end{figure*}

Most of the data-dependent regularizations are  motivated  by the empirical observation that data of interest typically lie close to a manifold, an assumption that has previously assisted machine learning tasks such as nonlinear embedding \cite{nonlinear_embedding}, semi-supervised labeling \cite{belkin2006manifold}, and multi-task classification \cite{evgeniou2005learning}. In the context of DNN,  data-dependent regularization techniques include the tangent distance algorithm \cite{Simard_1993,Simard_2012}, tangent prop algorithm \cite{Simard_1992}, and manifold tangent classifier \cite{Rifai_2011}. Typically, these algorithms only focus on the geometry of the input data, and do not encourage the network to produce geometrically meaningful features. Moreover, it is typically hard to explicity parametrize the underlying manifold, and some of the algorithms require human input of the tangent planes or tangent vectors \cite{Goodfellow-et-al-2016}.

Motivated by the same manifold observation, we propose a new network regularization technique that focuses on the geometry of both the input data and the output features, and build low-dimensional-manifold-regularized neural networks (\textit{LDMNet}). This is inspired by a recent algorithm in image  processing, the \textit{low dimensional manifold model} \cite{ldmm,ldmm_wgl,ldmm_scientific}. The idea of \textit{LDMNet} is that the concatenation $(\bx_i,\bxi_i)$ of the input data $\bx_i$ and the output features $\bxi_i$ should sample a collection of low dimensional manifolds. This idea is loosely related to the Gaussian mixture model: instead of assuming that the generating distribution of the input data is a mixture of Gaussians, we assume that the input-feature tuples $(\bx_i,\bxi_i)$ are generated  by a mixture of low dimensional manifolds. To emphasize this, we explicitly penalize the loss function with a term using an elegant formula from differential geometry to compute the dimension of the underlying manifold. The resulting variational problem is then solved via alternating minimization with respect to the manifold and the network weights. The corresponding Euler-Lagrange equation is a Laplace-Beltrami equation on a point cloud, which is efficiently solved by the point integral method (PIM) \cite{pim} with $O(N)$ computational complexity, where $N$ is the size of the input data. In \textit{LDMNet}, we never have to explicitly parametrize the manifolds or derive the tangent planes, and the solution is obtained by solving only the variational problem.

In the experiments, we demonstrate two benefits of \textit{LDMNet}: First, by extracting geometrically meaningful features, \textit{LDMNet}  significantly outperforms widely-used regularization techniques such as weight decay and \textit{DropOut}. For example, Figure~\ref{fig:visual_mnist} shows the two-dimensional projections of the test data from MNIST and their features learned from 1,000 training data by the same network with different regularizers. It can be observed that the features learned by weight decay and \textit{DropOut} typically sample two-dimensional regions, whereas the features learned by \textit{LDMNet} tend to lie close to one-dimensional and zero-dimensional manifolds (curves and points). Second, in some imaging problems, \textit{LDMNet} is more likely to find the model that is subject to different illumination patterns. By regularizing the network outputs, \textit{LDMNet}  can extract common features of the same subject imaged via different modalities so that these features sample the same low dimensional manifold. This can be observed in Figure~\ref{fig:ldm}, where the features of the same subject  extracted by \textit{LDMNet} from visible  (VIS) spectrum images and  near-infrared (NIR) spectrum images merge to form a single low dimensional manifold. This significantly increases the accuracy of cross-modality face recognition. The details of the experiments will be explained in Section \ref{sec:experiments}.

\section{Model Formulation}

For simplicity of explanation, we consider a $K$-way classification using a DNN. Assume that $\{(\bx_i,y_i)\}_{i=1}^N\subset \R^{d_1} \times \{1,\ldots,K\}$ is the labeled training set, and $\btheta$ is the collection of network weights. For every datum $\bx_i$ with class label $y_i \in \{1,\ldots,K\}$, the network first learns a $d_2$-dimensional feature $\bxi_i=f_{\btheta}(\bx_i)\in\R^{d_2}$, and then applies a softmax classifier to obtain the probability distribution of $\bx_i$ over the $K$ classes. The softmax loss $\ell(f_{\btheta}(\bx_i),y_i)$ is then calculated for $\bx_i$ as a negative log-probability of class $y_i$. The empirical loss function $J(\btheta)$ is defined as the average loss on the training set:
\begin{align}
  \label{eq:loss}
  J(\btheta) = \frac{1}{N}\sum_{i=1}^N\ell(f_{\btheta}(\bx_i),y_i).
\end{align}
When the training samples are scarce, statistical learning theories predict that overfitting to the training data  will occur \cite{Vapnik_1999}. What this means is that the average loss on the testing set can still be large even if the empirical loss $J(\btheta)$ is trained to be small.

\textit{LDMNet} provides an explanation and a solution for network overfitting. Empirical observation suggests that many data of interest typically sample a collection of low dimensional manifolds, i.e. $\{\bx_i\}_{i=1}^N \subset \N =\cup_{l=1}^L\N_l \subset \R^{d_1}$. One would also expect that  the feature extractor, $f_{\btheta}$, of a good learning algorithm  be a smooth function over $\N$ so that small variation in $\bx \in \N$ would not lead to dramatic change in the learned feature $\bxi = f_{\btheta}(\bx)\in \R^{d_2}$. Therefore the concatenation of the input data and output features,  $\{(\bx_i,\bxi_i)\}_{i=1}^N$, should sample a collection of low dimensional manifolds $\M = \cup_{l=1}^L\M_l \subset \R^{d}$, where $d = d_1+d_2$, and $\M_l=\left\{(\bx, f_{\btheta}(\bx))\right\}_{x \in \N_l}$ is the graph of $f_{\btheta}$ over $\N_l$. We suggest that network overfitting occurs when $\dim(\M_l)$ is too large after training. Therefore, to reduce overfitting,  we explicitly use the dimensions of $\M_l$ as a regularizer in the following variational form:
\begin{align}
  \label{eq:ldm}
 &\min_{\btheta, \M} \quad J(\btheta) + \frac{\lambda}{|\M|} \int_{\M}\dim(\M(\bm{p}))d\bm{p}\\ \nonumber
& \text{s.t. } \quad \{(\bx_i,f_{\btheta}(\bx_i))\}_{i=1}^N \subset \M,
\end{align}
where for any $\bm{p} \in \M = \cup_{l=1}^L\M_l$, $\M(\bm{p})$ denotes the manifold $\M_l$ to which $\bm{p}$ belongs, and $|\M| = \sum_{l=1}^L|\M_l|$ is the volume of $\M$.
The following theorem from differential geometry provides an elegant way of calculating the manifold dimension in (\ref{eq:ldm}).
\begin{thm} 
\label{thm:dim}
{\normalfont{\cite{ldmm}}} Let $\M$ be a smooth submanifold isometrically embedded in $\R^d$. For any $\bm{p}=(p_i)_{i=1}^d\in \M$,  
  \begin{align*}
    \dim(\M)=\sum_{i=1}^d\left|\nabla_\M\alpha_i(\bm{p})\right|^2,
  \end{align*}
where $\alpha_i(\bm{p})=p_i$ is the coordinate function, and $\nabla_\M$ is the gradient operator on the manifold $\M$. More specifically, $\nabla_\M \alpha_i = \sum_{s,t=1}^{k}g^{st}\partial_t\alpha_i\partial_s$, where $k$ is the intrinsic dimension of $\M$, and $g^{st}$ is the inverse of the metric tensor.
\end{thm}

As a result of Theorem \ref{thm:dim}, (\ref{eq:ldm}) can be reformulated as:
\begin{align}
  \label{eq:energy_alpha}
  &\min_{\btheta, \M} \quad J(\btheta) + \frac{\lambda}{|\M|} \sum_{j=1}^d \|\nabla_\M\alpha_j\|_{L^2(\M)}^2 \\ \nonumber
  &\text{s.t. } \quad \{(\bx_i,f_{\btheta}(\bx_i))\}_{i=1}^N \subset \M
\end{align}
where $\sum_{j=1}^d \|\nabla_\M\alpha_j\|_{L^2(\M)}^2$ corresponds to the $L^1$ norm of the local dimension. To solve (\ref{eq:energy_alpha}), we alternate the direction of minimization with respect to $\M$ and $\btheta$. More specifically, given $(\btheta^{(k)},\M^{(k)})$ at step $k$ satisfying $\{(\bx_i,f_{\btheta^{(k)}}(\bx_i))\}_{i=1}^N \subset \M^{(k)}$, step $k+1$ consists of the following
\begin{itemize}
\item Update $\btheta^{(k+1)}$ and the perturbed coordinate functions $\bm{\alpha}^{(k+1)}=(\alpha_1^{(k+1)},\cdots,\alpha_d^{(k+1)})$ as the minimizers of (\ref{eq:update_two}) with the fixed manifold $\M^{(k)}$:
  \begin{align}\label{eq:update_two}
    &\min_{\btheta, \bm{\alpha}} \quad J(\btheta)+\frac{\lambda}{|\M^{(k)}|} \sum_{j=1}^d \|\nabla_{\M^{(k)}}\alpha_j\|_{L^2(\M^{(k)})}^2\\ \nonumber
    & \text{s.t. } \quad \bm{\alpha}(\bx_i,f_{\btheta^{(k)}}(\bx_i)) = (\bx_i,f_{\btheta}(\bx_i)),\quad \forall i= 1,\ldots, N
  \end{align}
\item Update $\M^{(k+1)}$:
  \begin{align}
    \label{eq:update_M}
    \M^{(k+1)} = \bm{\alpha}^{(k+1)}(\M^{(k)})
  \end{align}
\end{itemize}
\begin{remark}\normalfont
As mentioned in Theorem \ref{thm:dim}, $\bm{\alpha}=(\alpha_1,\cdots,\alpha_d)$ is supposed to be the coordinate functions. In (\ref{eq:update_two}), we solve a perturbed version $\bm{\alpha}^{(k+1)}$ so that it maps the previous iterates of point cloud $\{(\bx_i,f_{\btheta^{(k)}}(\bx_i))\}_{i=1}^N$ and  manifold $\M^{(k)}$ to their corresponding updated versions. If the iteration converges to a fixed point, the consecutive iterates of the manifolds $\M^{(k+1)}$ and $\M^{(k)}$ will be very close to each other for sufficiently large $k$, and $\bm{\alpha}^{(k+1)}$ will be very close to the coordinate functions.
\end{remark}

Note that (\ref{eq:update_M}) is straightforward to implement, and (\ref{eq:update_two}) is an optimization problem with linear constraint, which can be solved via the alternating direction method of multipliers (ADMM). More specifically,
\begin{align}
  \nonumber
  & \bm{\alpha_{\bxi}}^{(k+1)} = \arg \min_{\bm{\alpha_{\bxi}}}  \sum_{j=d_1+1}^d\|\nabla_{\M^{(k)}}\alpha_j\|_{L^2(\M^{(k)})}\\    \label{eq:update_alpha}
 &+ \frac{\mu|\M^{(k)}|}{2\lambda N}\sum_{i=1}^N\|\bm{\alpha}_{\bxi}(\bx_i,f_{\btheta^{(k)}}(\bx_i))-(f_{\btheta^{(k)}}(\bx_i)-Z_i^{(k)})\|^2_2.
\end{align}
\begin{align}
  \nonumber
  \btheta^{(k+1)} = &\arg \min_{\btheta} J(\btheta)+ \frac{\mu}{2N}\sum_{i=1}^N\|\bm{\alpha}_{\bxi}^{(k+1)}(\bx_i,f_{\btheta^{(k)}}(\bx_i))\\ \label{eq:update_theta}
&-(f_{\btheta}(\bx_i)-Z_i^{(k)})\|^2_2.
\end{align}
\begin{align}
  \label{eq:update_z}
  Z_i^{(k+1)} = Z_i^{(k)} + \bm{\alpha}_{\bxi}^{(k+1)}(\bx_i,f_{\btheta^{(k)}}(\bx_i))-f_{\btheta^{(k+1)}}(\bx_i),
\end{align}
where $\bm{\alpha_\bxi}$ is defined as $\bm{\alpha} = (\bm{\alpha_{\bx}},\bm{\alpha_{\bxi}}) = \left((\alpha_1,\ldots,\alpha_{d_1}),(\alpha_{d_1+1},\ldots,\alpha_{d})\right)$, and $Z_i$ is the dual variable. Note that we  need to perturb only the coordinate functions $\bm{\alpha_{\bxi}}$ corresponding to the features in (\ref{eq:update_alpha}) because the inputs $\bx_i$ are given and fixed. Also note that because the gradient and the $L^2$ norm in (\ref{eq:update_alpha}) are defined on $\M$ instead of the projected manifold, we are \textit{not} simply minimizing the dimension of the manifold $\M$ projected onto the feature space. For computational efficiency, we update $\bm{\alpha},\btheta$ and $Z_i$ only once every manifold update (\ref{eq:update_M}).

Among (\ref{eq:update_alpha}),(\ref{eq:update_theta}) and (\ref{eq:update_z}), (\ref{eq:update_z}) is the easiest to implement, (\ref{eq:update_theta}) can be solved by stochastic gradient descent (SGD) with modified back propagation, and (\ref{eq:update_alpha}) can be solved by the point integral method (PIM) \cite{pim}. The detailed implementation of (\ref{eq:update_alpha}) and (\ref{eq:update_theta}) will be explained in the next section.

\section{Implementation Details and Complexity Analysis}

In this section, we present the details of the algorithmic implementation, which includes  back propagation for the $\btheta$ update (\ref{eq:update_theta}), point integral method for the $\bm{\alpha}$ update (\ref{eq:update_alpha}), and the complexity analysis.
\subsection{Back Propagation for the $\btheta$ Update}
We derive the gradient of the objective function in (\ref{eq:update_theta}). Let
\begin{align}
  \label{eq:second_term}
  E_i(\btheta)=\frac{\mu}{2}\|\bm{\alpha}_{\bxi}^{(k+1)}(\bx_i,f_{\btheta^{(k)}}(\bx_i))-(f_{\btheta}(\bx_i)-Z_i^{(k)})\|^2_2.
\end{align}
Then the objective function in (\ref{eq:update_theta}) is
\begin{align}
  \label{eq:loss_augmented}
  \tilde{J}(\btheta) = \frac{1}{N}\sum_{i=1}^N\ell(f_{\btheta}(\bx_i),y_i)+\frac{1}{N}\sum_{i=1}^NE_i(\btheta).
\end{align}
Usually the back-propagation of the first term in (\ref{eq:update_theta}) is known for a given network. As for the second term, let $\bx_i$ be a given datum in a mini-batch. The gradient of the second term with respect to the output layer $f_{\btheta}(\bx_i)$ is:
\begin{align}
  \label{eq:gradient_E}
  \frac{\partial E_i}{\partial f_{\btheta}(\bx_i)} = \mu\left(f_{\btheta}(\bx_i)-Z_i^{(k)}-\bm{\alpha_{\bxi}}^{(k+1)}(\bx_i,f_{\btheta^{(k)}}(\bx_i))\right)
\end{align}
This means that we need to only add the extra term (\ref{eq:gradient_E}) to the original gradient, and then use the already known procedure to back-propagate the gradient. This essentially leads to no extra computational cost in the SGD updates.

\subsection{Point Integral Method for the $\bm{\alpha}$ Update}

Note that the objective funtion in (\ref{eq:update_alpha}) is decoupled with respect to $j$, and each $\alpha_j$ update can be cast into:
\begin{align}
  \label{eq:cannonical}
  \min_{u \in H^1(\M)} \|\nabla_\M u\|_{L^2(\M)}^2 + \gamma \sum_{\bq \in P} |u(\bq) - v(\bq)|^2,
\end{align}
where $u = \alpha_j$, $\M = \M^{(k)}$, $\gamma = \mu |\M^{(k)}|/2\lambda N$, and $P = \left\{\bp_i=\left(\bx_i,f_{\btheta^{(k)}}(\bx_i)\right)\right\}_{i=1}^N\subset \M$. The Euler-Lagrange equation of (\ref{eq:cannonical}) is:
\begin{align}\nonumber
  -\Delta_\M u(\bp) + \gamma \sum_{\bq\in P}\delta(\bp-\bq)(u(\bq)-v(\bq))&= 0, \text{ } \bp \in \M\\ \label{eq:EL}
\frac{\partial u}{\partial n} &= 0, \text{ } \bp \in \partial \M
\end{align}
It is hard to discretize the Laplace-Beltrami operator $\Delta_\M$ and the delta function $\delta(x,y)$ on an unstructured point cloud $P$. We instead use the point integral method to solve (\ref{eq:EL}). The key observation in PIM is the following theorem:

\begin{thm} {\normalfont\cite{pim}}
  \label{thm:local-error}
If $u\in C^3(\M)$ is a function on $\M$,
then
\begin{align}\nonumber
  &\left\|\int_\M \Delta_{\M} u(\bq)R_t(\bp,\bq)d \bq- 2\int_{\partial \M} \frac{\partial u(\bq)}{\partial n} R_t(\bp, \bq) d \tau_{\bq}\right.\\
  & +\left.\frac{1}{t}\int_{\M} (u(\bp)-u(\bq))R_t(\bp, \bq) d \bq\right\|_{L^2(\M)}=O(t^{1/4}).
\end{align}
where $R_t$ is the normalized heat kernel:
\begin{align}
  \label{eq:Rt}
  R_t(\bp,\bq) = C_t\exp\left(-\frac{|\bp-\bq|^2}{4t}\right).
\end{align}
\end{thm}

After convolving equation (\ref{eq:EL}) with the heat kernel $R_t$, we know the solution $u$ of (\ref{eq:EL}) should satisfy
\begin{align}
  \nonumber
  &-\int_\M \Delta_{\M} u (\bm{q})R_t(\bm{p},\bm{q})d\bm{q}\\  \label{eq:EL-convolve}
  + &\gamma\sum_{\bm{q}\in P}R_t(\bm{p},\bm{q})\left(u(\bm{q})-v(\bm{q})\right) = 0.
\end{align}
Combined with Theorem \ref{thm:local-error} and the  Neumann boundary condition, this implies that $u$ should approximately satisfy
\begin{align}
 \nonumber
  &\int_\M\left(u(\bp)-u(\bq)\right)R_t(\bp,\bq)d\bq\\
\label{eq:linear-system-continuous}
+ &\gamma t \sum_{\bq \in P}R_t(\bp,\bq)\left(u(\bq) - v(\bq)\right) = 0
\end{align}

Note that (\ref{eq:linear-system-continuous}) no longer involves the gradient $\nabla_\M$ or the Laplace-Beltrami operator $\Delta_\M$. Assume that $P = \{\bp_1,\ldots,\bp_N\}$ samples the manifold $\M$ uniformly at random, then (\ref{eq:linear-system-continuous}) can be discretized as
\begin{align}
  \label{eq:linear-system-discretize}
\frac{|\M|}{N}\sum_{j=1}^NR_{t,ij}(u_i-u_j) +\gamma t \sum_{j=1}^N R_{t,ij} (u_j - v_j) = 0,
\end{align}
where $u_i= u(\bp_i)$, and $R_{t,ij}= R_t(\bp_i,\bp_j)$. Combining the definition of $\gamma$ in (\ref{eq:cannonical}), we can write (\ref{eq:linear-system-discretize}) in the matrix form
\begin{align}
  \label{eq:linear-system-matrix}
\left(\bm{L}+  \frac{\mu}{\tilde{\lambda}} \bm{W}\right) \bm{u} = \frac{\mu}{\tilde{\lambda}} \bm{W} \bm{v}, \quad \tilde{\lambda} = 2\lambda/t,
\end{align}
where $\tilde{\lambda}$ can be chosen instead of $\lambda$ as the hyperparameter to be tuned, $\bm{u} = (u_1, \ldots, u_N)^T$, $\bm{W}$ is an $N\times N$ matrix
\begin{align}
  \label{eq:W}
  \bm{W}_{ij} = R_{t,ij} = \exp\left(-\frac{|\bp_i-\bp_j|^2}{4t}\right),
\end{align}
and $\bm{L}$ is the graph Laplacian of $\bm{W}$:
\begin{align}
  \label{eq:L}
  \bm{L}_{ii} = \sum_{j\not=i}\bm{W}_{ij}, \quad \text{and}\quad \bm{L}_{ij} = -\bm{W}_{ij}\quad \text{if} \quad i\not=j.
\end{align}
Therefore, the update of $\bm{\alpha_\bxi}$, which is cast into the canonical form (\ref{eq:cannonical}), is achieved by solving a linear system (\ref{eq:linear-system-matrix}).

\subsection{Complexity Analysis}
\label{sec:complexity}
Based on the analysis above, we present a summary of the traning for \textit{LDMNet} in Algorithm \ref{alg:ldmnet}. 
\begin{algorithm}
\floatname{algorithm}{Algorithm}
\caption{\textit{LDMNet} Training}
\label{alg:ldmnet}
\begin{algorithmic}
\REQUIRE Training data $\{(\bx_i,y_i)\}_{i=1}^N \subset \R^{d_1}\times \R$, hyperparameters $\tilde{\lambda}$ and $\mu$, and a neural network with the weights $\btheta$ and the output layer $\bxi_i = f_{\btheta}(\bx_i)\in \R^{d_2}$.
\ENSURE  Trained network weights $\btheta^*$.
\STATE Randomly initialize the network weights $\btheta^{(0)}$. The dual variables $Z_i^{(0)} \in \R^{d_2}$ are initialized to zero.
\WHILE {not converge}
\STATE 1. Compute the matrices $\bm{W}$ and $\bm{L}$ as in (\ref{eq:W}) and (\ref{eq:L}) with $\bm{p}_i = (\bx_i,f_{\btheta^{(k)}}(\bx_i))$.
\vspace{1mm}
\STATE 2. Update $\bm{\alpha}^{(k+1)}$ in (\ref{eq:update_alpha}): solve the linear systems (\ref{eq:linear-system-matrix}), where
\vspace{-3mm}
\begin{align}
  \label{eq:u_and_v}
  \bm{u}_i=\alpha_j(\bm{p}_i),\quad \bm{v}_i = f_{\btheta^{(k)}}(\bx_i)_j-Z_{i,j}^{(k)}.
\end{align}
\vspace{-5mm}
\STATE 3. Update $\btheta^{(k+1)}$ in (\ref{eq:update_theta}): run SGD for $M$ epochs with an extra gradient term (\ref{eq:gradient_E}).
\STATE 4. Update $Z^{(k+1)}$ in (\ref{eq:update_z}).
\STATE 5. $k\leftarrow k+1$.
\ENDWHILE
\STATE $\btheta^*\leftarrow \btheta^{(k)}$.
\end{algorithmic}
\end{algorithm}

The additional computation in Algorithm \ref{alg:ldmnet} (in steps 1 and 2) comes from the update of weight matrices in (\ref{eq:W}) and solving the linear system (\ref{eq:linear-system-matrix}) from PIM once every $M$ epochs of SGD. We now explain the computational complexity of these two steps.

When $N$ is large, it is not computationally feasible to compute the pairwise distances in the entire training set. Therefore the weight matrix $\bm{W}$ is truncated to only $20$ nearest neighbors. To identify those nearest neighbors, we first organize the data points $\{\bm{p}_1,\ldots,\bm{p}_N\}\subset \R^d$ into a $k$-d tree \cite{kdtree}, which is a binary tree that recursively partitions a $k$-dimensional space (in our case $k = d$). Nearest neighbors can then be efficiently identified because branches can be eliminated from the search space quickly. Modern algorithms to build a balanced $k$-d tree generally at worst converge in $O(N\log N)$ time \cite{kdtree_build_1,kdtree_build_2}, and finding nearest neighbours for one query point in a balanced $k$-d tree takes $O(\log N)$ time on average \cite{kdtree_search}. Therefore the complexity of the weight update is $O(N\log N)$.

Since $\bm{W}$ and $\bm{L}$ are sparse symmetric matrices with a fixed maximum number of non-zero entries in each row, the linear system (\ref{eq:linear-system-matrix}) can be solved efficiently with the preconditioned conjugate gradients method. After restricting the number of matrix multiplications to a maximum of 50, the complexity of the $\bm{\alpha}$ update is $O(N)$.

\section{Experiments}
\label{sec:experiments}
In this section, we compare the performance of \textit{LDMNet} to widely-used network regularization techniques, weight decay and \textit{DropOut}, using the same underlying network structure. We point out that our focus is to compare the effectiveness of the regularizers, and not to investigate the state-of-the-art performances on the benchmark datasets. Therefore we typically use simple network structures and relatively small training sets, and no data augmentation or early stopping is implemented.

Unless otherwise stated, all experiments use mini-batch SGD with momentum on batches of 100 images. The momentum parameter is fixed at $0.9$. The networks are trained using a fixed learning rate $r_0$ on the first $200$ epochs, and then $r_0/10$ for another $100$ epochs.

As mentioned in Section \ref{sec:complexity}, the weight matrices $\bm{W}$ are truncated to $20$ nearest neighbors. For classification tasks, nearest neighbors can be searched within each class in the labeled training set. We also normalize the weight matrices with local scaling factors $\sigma(\bm{p})$ \cite{zelnik2005self}:
\begin{align}
  \label{eq:weight}
  w(\bm{p},\bm{q}) = \exp \left(-\frac{\|\bm{p}-\bm{q}\|^2}{\sigma(\bm{p})\sigma(\bm{q})}\right),
\end{align}
where $\sigma(\bm{p})$ is chosen as the distance between $\bm{p}$ and its $10$th nearest neighbor. This is based on the empirical analysis on choosing the parameter $t$ in \cite{pim}, and has been used in \cite{ldmm_scientific}. The weight matrices and $\bm{\alpha}$ are updated once every $M=2$ epochs of SGD.

All hyperparameters are optimized so that those reported are the best performance of each method. For \textit{LDMNet}, $\tilde{\lambda}$ defined in (\ref{eq:linear-system-matrix}) typically decreases as the training set becomes larger, whereas the paramter $\mu$ for augmented Lagrangian can be fixed to be a constant. For weight decay,
\begin{align}
  \label{eq:weight_decay}
  \min_{\btheta} J(\btheta) + w\|\btheta\|_2^2,
\end{align}
the parameter $w$ also usually decreases as the training size increases. For \textit{DropOut}, the corresponding \textit{DropOut} layer is always chosen to have a drop rate of $0.5$.

\subsection{MNIST}

The MNIST handwritten digit dataset contains approximately 60,000 training images ($28\times 28$) and 10,000 test images. Tabel \ref{tab:mnist_net} describes the network structure. The learned feature $f_{\btheta}(\bx_i)$ for the training data $\bx_i$ is the output of layer six, and is regularized for \textit{LDMNet} in (\ref{eq:ldm}).

\begin{table}
  \centering
  \begin{tabular}{c|c|C{4cm}}
    \hline\hline
    Layer& Type & Parameters\\
    \hline
    1 & conv & size: $5\times 5\times 1\times 20$ \newline stride: 1, pad: 0\\    
    \hline
    2 & max pool & size: $2\times 2$, stride: 2, pad: 0\\
    \hline
    3 & conv & size: $5\times 5\times 20\times 50$ \newline stride: 1, pad: 0\\
    \hline
    4 & max pool & size: $2\times 2$, stride: 2, pad: 0\\
    \hline
    5 & conv & size: $4\times 4\times 50\times 500$ \newline stride: 1, pad: 0\\
    \hline
    6 & ReLu (\textit{DropOut}) & N/A\\
    \hline
    7 & fully connected & $500 \times 10$\\
    \hline
    8 & softmaxloss & N/A\\
    \hline\hline
  \end{tabular}
  \vspace{1mm}
  \caption{Network structure in the MNIST experiments. The outputs of layer 6 are the extracted features, which will be fed into the softmax classifier (layer 7 and 8).}
  \label{tab:mnist_net}
\end{table}

\begin{table}
  \centering
  \begin{tabular}{C{2cm}|C{1.5cm}|C{1.5cm}|C{1.5cm}}
    \hline\hline
    training per class & $\tilde{\lambda}$ & $\mu$ & $w$ \\
    \hline
    50 & 0.05 & 0.01 & 0.1\\
    \hline
    100 & 0.05 & 0.01 & 0.05\\
    \hline
    400 & 0.01 & 0.01 & 0.01\\
    \hline
    700 & 0.01 & 0.01 & 0.005\\
    \hline
    1000 & 0.005 & 0.01 & 0.005\\
    \hline
    3000 & 0.001 & 0.01 & 0.001\\
    \hline
    6000 & 0.001 & 0.01 & 0.001\\
    \hline\hline
  \end{tabular}
  \vspace{1mm}
  \caption{Hyperparamters used in the MNIST experiments}
  \label{tab:mnist_hyper}
\end{table}

\begin{table}
  \centering
  \begin{tabular}{C{2cm}|C{1.5cm}|C{1.5cm}|C{1.5cm}}
    \hline\hline
    training per class & weight decay & \textit{DropOut} & \textit{LDMNet} \\
    \hline
    50 & 91.32\% & 92.31\% & \textbf{95.57\%}\\
    \hline
    100 & 93.38\% & 94.05\% & \textbf{96.73\%}\\
    \hline
    400 & 97.23\% & 97.95\% &\textbf{98.41\%} \\
    \hline
    700 & 97.67\% & 98.07\% & \textbf{98.61\%}\\
    \hline
    1000 &98.06\%  & 98.71\% & \textbf{98.89\%}\\
    \hline
    3000 & 98.87\% & 99.21\% & \textbf{99.24\%}\\
    \hline
    6000 & 99.15\% & \textbf{99.41\%} & 99.39\%\\
    \hline\hline
  \end{tabular}
  \vspace{1mm}
  \caption{MNIST: testing accuracy for different regularizers}
  \label{tab:mnist_result}
\end{table}

\begin{figure}[t]
  \centering
  \begin{tabular}{cc}
    \includegraphics[width=0.48\linewidth]{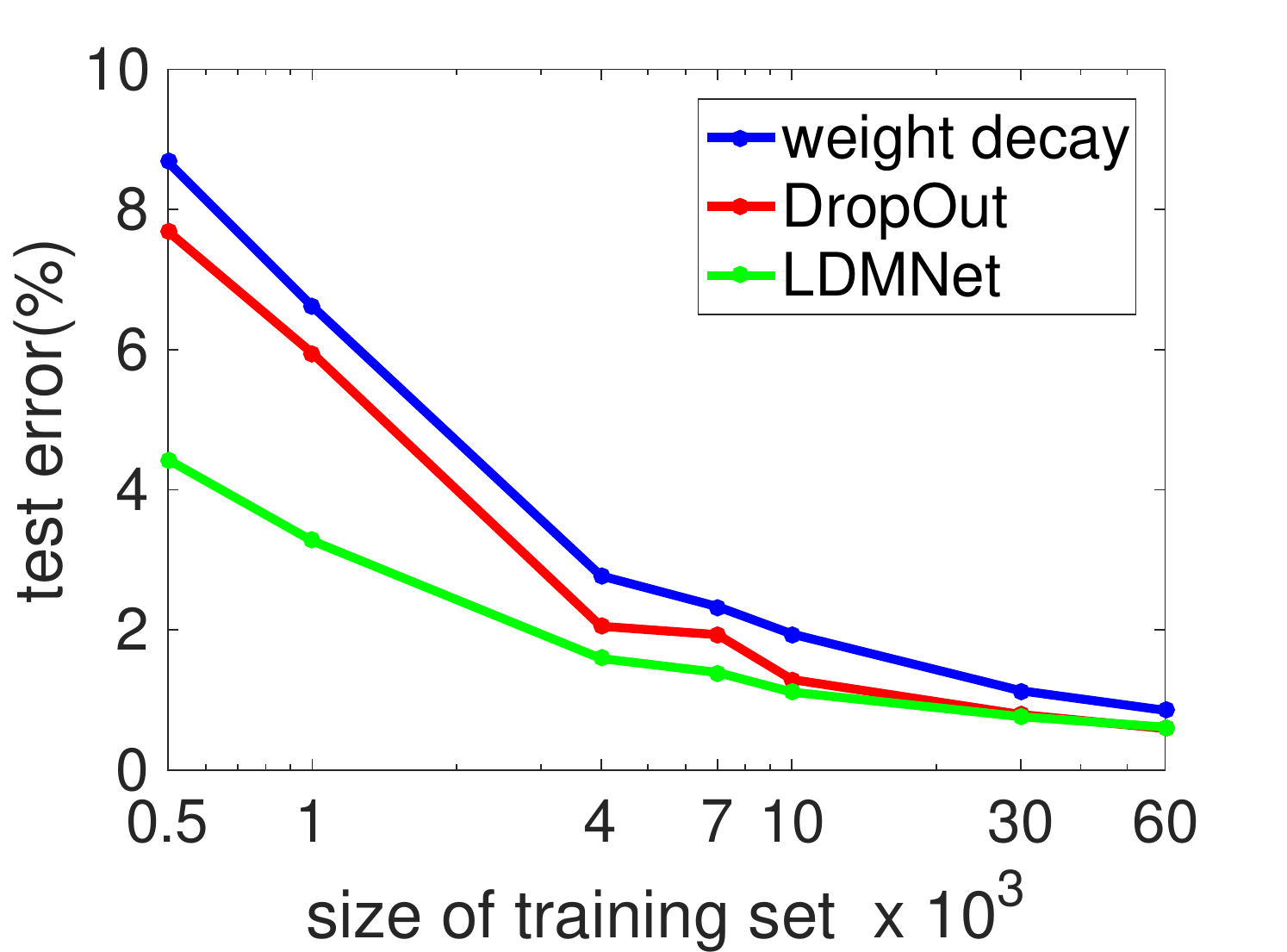}&
    \includegraphics[width=0.48\linewidth]{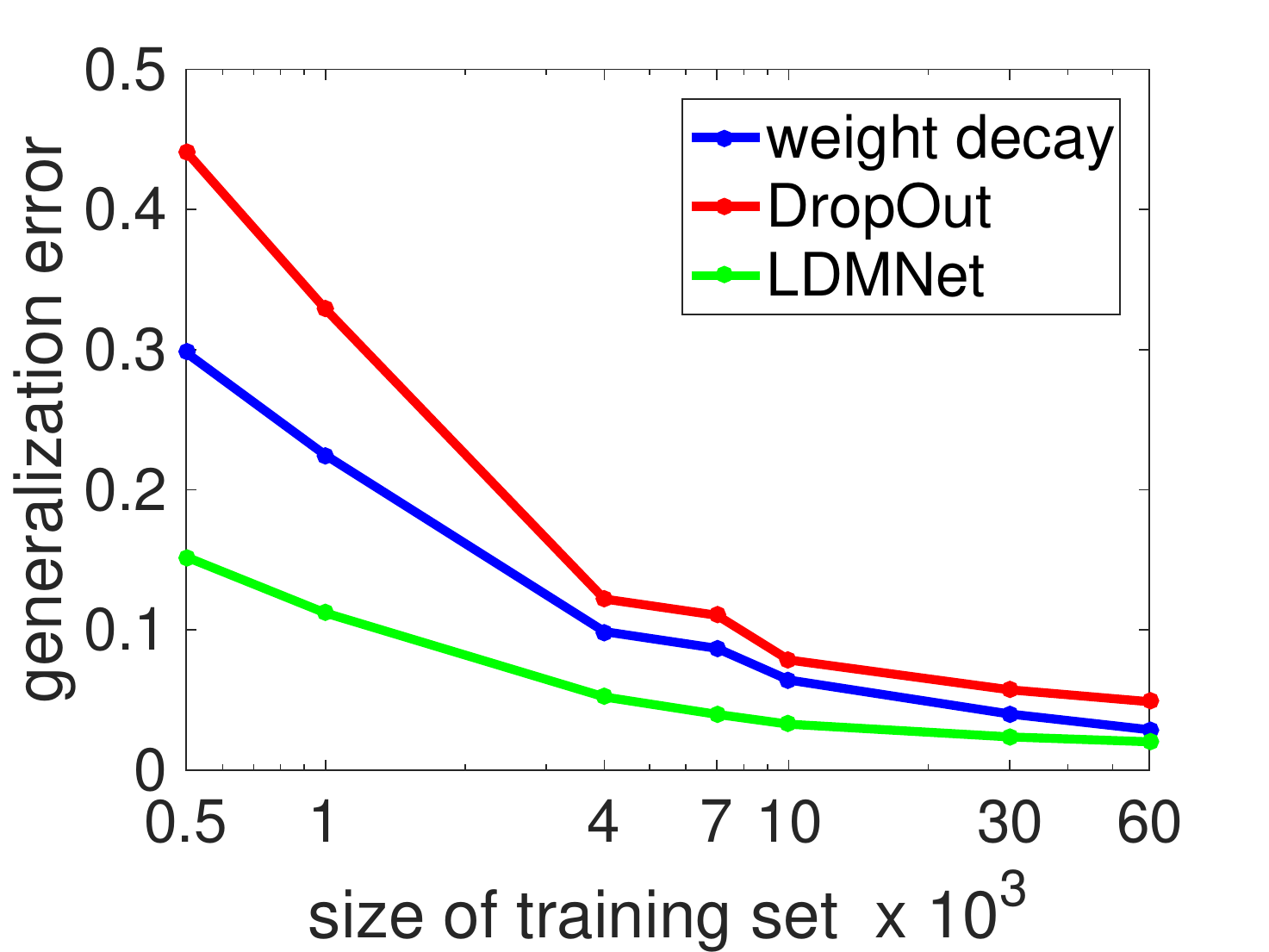}\\
  \end{tabular}
\caption{Comparison of the regularizers on the MNIST dataset. The first (second) figure shows the dependence of the classification (generalization) error on the size of the training set.}
\label{fig:mnist_error}
\end{figure}

While state-of-the-art methods often use the entire training set, we are interested in examining the performance of the regularization techniques with varying training sizes from 500 to 60,000. In this experiment, the initial learning rate is set to $0.001$, and the hyperparameters are reported in Table \ref{tab:mnist_hyper}. Table \ref{tab:mnist_result} displays the testing accuracy of the competing algorithms. The dependence of the classification error and generalization error (which is the difference between the softmax loss on the testing and training data) on the size of the training set is shown in Figure~\ref{fig:mnist_error}. Figure~\ref{fig:visual_mnist} provides a visual illustration of the features of the testing data learned from 1,000 training samples. It is clear to see that \textit{LDMNet} significantly outperforms weight decay and \textit{DropOut} when the training set is small, and the performance becomes broadly similar as the size of the training set reaches 60,000.

\subsection{SVHN and CIFAR-10}
\begin{table}
  \centering
  \begin{tabular}{c|c|C{4cm}}
    \hline\hline
    Layer& Type & Parameters\\
    \hline
    1 & conv & size: $5\times 5\times 3\times 96$ \newline stride: 1, pad: 2\\    
    \hline
    2 & ReLu & N/A\\
    \hline
    3 & max pool & size: $3\times 3$, stride: 2, pad: 0\\
    \hline
    4 & conv & size: $5\times 5\times 96\times 128$ \newline stride: 1, pad: 2\\
    \hline
    5 & ReLu & N/A\\
    \hline
    6 & max pool & size: $3\times 3$, stride: 2, pad: 0\\
    \hline
    7 & conv & size: $4\times 4\times 128\times 256$ \newline stride: 1, pad: 0\\
    \hline
    8 & ReLu & N/A\\
    \hline
    9 & max pool & size: $3\times 3$, stride: 2, pad: 0\\
    \hline
    10 & fully connected & output: 2048\\
    \hline
    11 & ReLu (\textit{DropOut}) & N/A \\
    \hline
    12 & fully connected & output: 2048\\
    \hline
    13 & ReLu (\textit{DropOut}) & N/A\\
    \hline
    14 & fully connected & $2048\times 10$\\
    \hline
    15 & softmaxloss & N/A\\
    \hline\hline
  \end{tabular}
  \vspace{1mm}
  \caption{Network structure in the SVHN and CIFAR-10 experiments. The outputs of layer 13 are the extracted features, which will be fed into the softmax classifier (layer 14 and 15).}
  \label{tab:svhn_cifar_net}
\end{table}

\begin{table}[!t]
  \centering
  \begin{tabular}{C{1.5cm}|c|C{1.5cm}|c|C{1.5cm}}
    \hline\hline
    training& \multicolumn{2}{c|}{SVHN}&\multicolumn{2}{c}{CIFAR-10}\\ 
    \cline{2-5}per class & $\tilde{\lambda}$ & $w$ &  $\tilde{\lambda}$ & $w$\\ \hline
    50  & 0.1  & $10^{-6}$ & 0.01 & $5\times 10^{-4}$\\ \hline
    100 & 0.05 & $10^{-6}$ & 0.01 & $5\times 10^{-5}$\\ \hline
    400 & 0.05 & $10^{-7}$ & 0.01 & $5\times 10^{-5}$\\ \hline
    700 & 0.01 & $10^{-8}$ & 0.01 & $5\times 10^{-7}$\\ \hline\hline
  \end{tabular}
  \vspace{1mm}
  \caption{Hyperparameters used in the SVHN and CIFAR-10 experiments.}
  \label{tab:svhn_cifar_hyper}
\end{table}

\begin{table}
  \centering
  \begin{tabular}{C{2cm}|C{1.5cm}|C{1.5cm}|C{1.5cm}}
    \hline\hline
    training per class & weight decay & \textit{DropOut} & \textit{LDMNet} \\
    \hline
    50 & 71.46\% & 71.94\% & \textbf{74.64\%}\\
    \hline
    100 & 79.05\% & 79.94\% & \textbf{81.36\%}\\
    \hline
    400 & 87.38\% & 87.16\% &\textbf{88.03\%} \\
    \hline
    700 & 89.69\% & 89.83\% & \textbf{90.07\%}\\
    \hline\hline
  \end{tabular}
  \vspace{1mm}
  \caption{SVHN: testing accuracy for different regularizers}
  \label{tab:svhn_result}
\end{table}

\begin{table}
  \centering
  \begin{tabular}{C{2cm}|C{1.5cm}|C{1.5cm}|C{1.5cm}}
    \hline\hline
    training per class & weight decay & \textit{DropOut} & \textit{LDMNet} \\
    \hline
    50 & 34.70\% & 35.94\% & \textbf{41.55\%}\\
    \hline
    100 & 42.45\% & 43.18\% & \textbf{48.73\%}\\
    \hline
    400 & 56.19\% & 56.79\% &\textbf{60.08\%} \\
    \hline
    700 & 61.84\% & 62.59\% & \textbf{65.59\%}\\
    \hline
    full data & \multicolumn{2}{c|}{87.72\%} & \textbf{88.21\%}\\
    \hline\hline
  \end{tabular}
  \vspace{1mm}
  \caption{CIFAR-10: testing accuracy for different regularizers. The first four experiments use training sets varied in size from 500 to 7,000 and a simple network specified in Table~\ref{tab:svhn_cifar_net}. In the last experiment, we first train a DNN with VGG-16 architecture \cite{vgg16} from the full training data using both weight decay and \textit{DropOut}, then the DNN is fine-tuned by regularizing the output layer with \textit{LDMNet}. No data augmentation is implemented in any of the experiment.}
  \label{tab:cifar_result}
\end{table}

SVHN and CIFAR-10 are benchmark RGB image datasets, both of which contain 10 different classes. These two datasets are  more challenging than the MNIST dataset because of their weaker intraclass correlation. All algorithms use the network structure (similar to \cite{hinton2012improving}) specified in Table~\ref{tab:svhn_cifar_net}. The outputs of layer 13 are the learned features, and are regularized in \textit{LDMNet}. All algorithms start with a learning rate of $0.005$ for SVHN and $0.001$ for CIFAR-10. $\mu$ has been fixed as $0.5$ for SVHN and $1$ for CIFAR-10, and the remaining hyperparameters are reported in Table~\ref{tab:svhn_cifar_hyper}.

We report the testing accuracies of the competing regularizers in Table~\ref{tab:svhn_result} and Table~\ref{tab:cifar_result} when the number of training samples is varied from $50$ to $700$ per class. Again, it is clear to see that \textit{LDMNet} outperforms weight decay and \textit{DropOut} by a significant margin.

To demonstrate the generality of \textit{LDMNet}, we conduct another experiment on CIFAR-10 using the entire training data with a different network structure. We first train a DNN with VGG-16 architecture \cite{vgg16} on CIFAR-10 using both weight decay and \textit{DropOut} without data augmentation. Then, we fine-tune the DNN by regularizing the output layer with \textit{LDMNet}. The testing accuracies are reported in the last row of Table~\ref{tab:cifar_result}. Again, \textit{LDMNet} outperforms weight decay and \textit{DropOut}, demonstrating that \textit{LDMNet} is a general framework that can be used to improve the performance of any network structure.

\subsection{NIR-VIS Heterogeneous Face Recognition}

\begin{figure}[t]
  \centering
  \begin{tabular}{cccc}
    \includegraphics[width=0.24\linewidth]{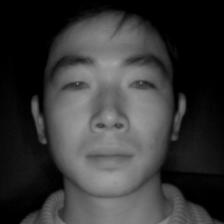}&\hspace{-4mm}
    \includegraphics[width=0.24\linewidth]{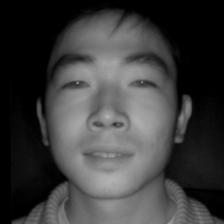}&\hspace{-4mm}
    \includegraphics[width=0.24\linewidth]{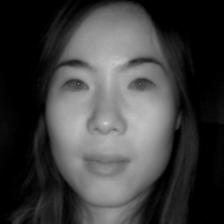}&\hspace{-4mm}
    \includegraphics[width=0.24\linewidth]{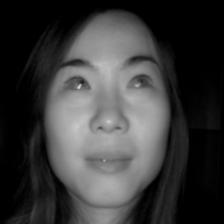}\\
    \includegraphics[width=0.24\linewidth]{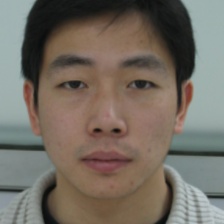}&\hspace{-4mm}
    \includegraphics[width=0.24\linewidth]{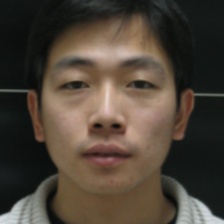}&\hspace{-4mm}
    \includegraphics[width=0.24\linewidth]{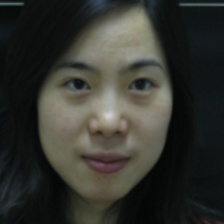}&\hspace{-4mm}
    \includegraphics[width=0.24\linewidth]{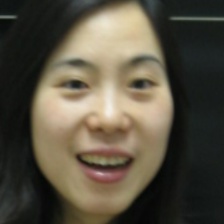}
  \end{tabular}
\caption{Sample images of two subjects from the CASIA NIR-VIS 2.0 dataset after the pre-procssing of alignment and cropping \cite{Kazemi_2014_CVPR}. Top: NIR. Bottom: VIS.}
\label{fig:nirvis_sample}
\end{figure}

Finally, we demonstrate the effectiveness of \textit{LDMNet} for NIR-VIS face recognition. The objective of the experiment is to match a probe image of a subject captured in the near-infrared spectrum (NIR) to the same subject from a gallery of visible spectrum (VIS) images. The CASIA NIR-VIS 2.0 benchmark dataset \cite{nirvis} is used to evaluate the performance. This dataset contains 17,580 NIR and VIS face images of 725 subjects. Figure~\ref{fig:nirvis_sample} shows eight sample images of two subjects after facial landmark alignment and cropping \cite{Kazemi_2014_CVPR}. Despite recent breakthroughs for VIS face recognition by training DNNs from millions of VIS images, such approach cannot be simply transferred to NIR-VIS face recognition. The reason is that, unlike VIS face images, we have only limited number of availabe NIR images. Moreover, the NIR-VIS face matching is a cross-modality comparison.

The authors in \cite{lezama2016not} introduced a way to transfer the breakthrough in VIS face recognition to the NIR spectrum. Their idea is to use a DNN pre-trained on VIS images as a feature extactor, while making two independent modifications in the input and output of  the DNN. They first modify the input by ``hallucinating'' a VIS image from the NIR sample, and then apply a low-rank embedding of the DNN features at the output. The combination of these two modifications achieves the state-of-the-art performance on cross-spectral face recognition.

\begin{table}
  \centering
  \begin{tabular}{c|c|C{3cm}}
    \hline\hline
    Layer& Type & Parameters\\
    \hline
    1 & fully connected & output:2000\\
    \hline
    2 & ReLu (\textit{DropOut})& N/A\\
    \hline
    3 & fully connected & output:2000\\
    \hline
    4 & ReLu (\textit{DropOut})& N/A\\
    \hline\hline
  \end{tabular}
  \vspace{1mm}
  \caption{Fully connected network for the NIR-VIS nonlinear feature embedding. The outputs of layer 4 are the extracted features.}
  \label{tab:nirvis_net}
\end{table}

We follow the second idea in \cite{lezama2016not}, and learn a nonlinear low dimensional manifold embedding of the output features. The intuition is that faces of the same subject in two different modalities should sample the same low dimensional feature manifold in a transformed space. In our experiment, we use the VGG-face model (downloaded at \url{http://www.robots.ox.ac.uk/~vgg/software/vgg_face/}) \cite{parkhi2015deep} as a feature extractor. The learned 4,096 dimensional features can be reduced to a 2,000 dimensional space using PCA and used directly for face matching. Meanwhile, we also put the 4,096 dimensional features into a two-layer fully connected network described in Table~\ref{tab:nirvis_net} to learn a nonlinear embedding using different regularizations. The features extracted from layer 4 are regularized in \textit{LDMNet}.

All nonlinear embeddings using the structure specified in Table~\ref{tab:nirvis_net} are trained with SGD on mini-batches of 100 images for 200 epochs. We use an exponentially decreasing  learning rate that starts at 0.1 with a decaying factor of 0.99. The hyperparameters are chosen to achieve the optimal performance on the validation set. More specifically, $\tilde{\lambda}, \mu, w$ are set to $5\times 10^{-5},5$, and $5\times 10^{-4}$ respectively.

\begin{table}
\vspace{1mm}
  \centering
  \begin{tabular}{lc}
    \hline
     & Accuracy (\%)\\
    \hline\hline
    VGG-face & $74.51 \pm 1.28$\\
    VGG-face + triplet \cite{lezama2016not}& $75.96 \pm 2.90$\\
    VGG-face + low-rank \cite{lezama2016not}& $80.69 \pm 1.02$\\
    VGG-face weight Decay & $63.87 \pm 1.33$\\
    VGG-face \textit{DropOut} & $66.97 \pm 1.31$\\
    VGG-face \textit{LDMNet} & \textbf{85.02} $\bm{\pm}$ \textbf{0.86}\\
    \hline\hline
  \end{tabular}
  \vspace{1mm}
  \caption{NIR-VIS cross-spectral rank-1 identification rate on the 10-fold CASIA NIR-VIS 2.0 benchmark. The first result is obtained by reducing the features learned from VGG-face to 2,000 dimensional space using PCA. The next two results use triplet \cite{weinberger2006distance} and low-rank embedding of the learned features, and are reported in \cite{lezama2016not}. The last three results are achieved by training the nonlinear embedding network in Table~\ref{tab:nirvis_net} with the corresponding regularizations.}
  \label{tab:nirvis_result}
\end{table}

We report the rank-1 performance score for the standard CASIA NIR-VIS 2.0 evaluation protocol in Table~\ref{tab:nirvis_result}. Because of the limited amount of training data (around 6,300 NIR and 2,500 VIS images), the fully-connected networks in Table~\ref{tab:nirvis_net} trained with weight decay and \textit{DropOut} clearly overfit the training data: they actually yield testing accuracies that are worse than using a simple PCA embedding of the features learned from VGG-face. However, the same network regularized with \textit{LDMNet} has achieved a significant 10.5\% accuracy boost (from 74.51\% to 85.02\%) to using VGG-face directly. It is also better than the results reported in \cite{lezama2016not} using the popular triplet embedding \cite{weinberger2006distance} and low-rank embedding. Figure~\ref{fig:visual} provides a visual illustration of the learned features from different regularizations. The generated features of five subjects are visualized in two dimensions using PCA, with filled circle for VIS, and unfilled diamond for NIR, and one color for each subject. Note that in Figure~\ref{fig:vgg},\ref{fig:wd},\ref{fig:dropout}, features of one subject learned directly from VGG-face or from a nonlinear embedding regularized with weight decay or \textit{DropOut} typically form two clusters, one for NIR and the other for VIS. This contrasts with the behavior in Figure~\ref{fig:ldm}, where features of the same subject from two different modalities merge to form a single low dimensional manifold.

\section{Conclusion}
We proposed a general deep neural network regularization technique \textit{LDMNet}. The intuition of \textit{LDMNet} is that the concatenation of the input data and output features should sample a collection of low dimensional manifolds, an idea that is loosely related to the Gaussian mixture model.  Unlike most data-dependent regularizations, \textit{LDMNet} focuses on the geometry of both the input data and the output features, and does not require explicit parametrization of  the underlying manifold. The dimensions of the manifolds are directly regularized in a variational form, which is solved by alternating direction of minimization with a slight increase in the computational complexity ($O(N)$ in solving the linear system, and $O(N\log N)$ in the weight update). Extensive experiments show that \textit{LDMNet} has two benefits: First, it significantly outperforms widely used regularization techniques, such as weight decay and \textit{DropOut}. Second, \textit{LDMNet} can extract common features  of the same subject imaged via different modalities so that these features sample the same low dimensional manifold, which significantly increases the accuracy of cross-spectral face recognition.

{\small
\bibliographystyle{abbrv}
\bibliography{ldmnet}
}

\end{document}